\documentclass[11pt]{article}
\usepackage{coling2018}
\usepackage{times}
\usepackage{url}
\usepackage{latexsym}

\usepackage{graphicx}
\usepackage{amsmath}
\usepackage{CJKutf8}
\usepackage{algorithm}  
\usepackage{algorithmic} 
\usepackage{multirow}

\title{Attribute Acquisition in Ontology based on Representation Learning of Hierarchical Classes and Attributes}

\author{Tianwen Jiang, Ming Liu, Bing Qin, Ting Liu \\
		Research Center for Social Computing and Information Retrieval \\ 
		Harbin Institute of Technology, China \\
		{\tt \{twjiang, mliu, bqin, mliu\}@ir.hit.edu.cn} \\}

\date{}

\begin{document}
\begin{CJK*}{UTF8}{gbsn} 

\maketitle
\begin{abstract}
\textit{Attribute acquisition} for classes is a key 
step in ontology construction, which is often achieved 
by community members manually. This paper investigates 
an attention-based automatic paradigm called \textbf{TransATT} 
for attribute acquisition, by learning the representation 
of hierarchical classes and attributes in Chinese ontology. 
The attributes of an entity can be acquired by merely 
inspecting its classes, because the entity can be regard 
as the instance of its classes and inherit their attributes. 
For explicitly describing of the class of an entity unambiguously, 
we propose \textit{class-path} to represent the hierarchical 
classes in ontology, instead of the terminal class word
of the hypernym-hyponym relation (i.e., \textit{is-a} relation) 
based hierarchy. The high performance of TransATT on attribute 
acquisition indicates the promising ability of the learned 
representation of class-paths and attributes. Moreover, 
we construct a dataset named \textbf{BigCilin11k}. To the 
best of our knowledge, this is the first Chinese dataset 
with abundant hierarchical classes and entities with attributes.
\end{abstract}

\section{Introduction}

\begin{figure}[htb]
	\begin{center}
		\includegraphics[width=10cm]{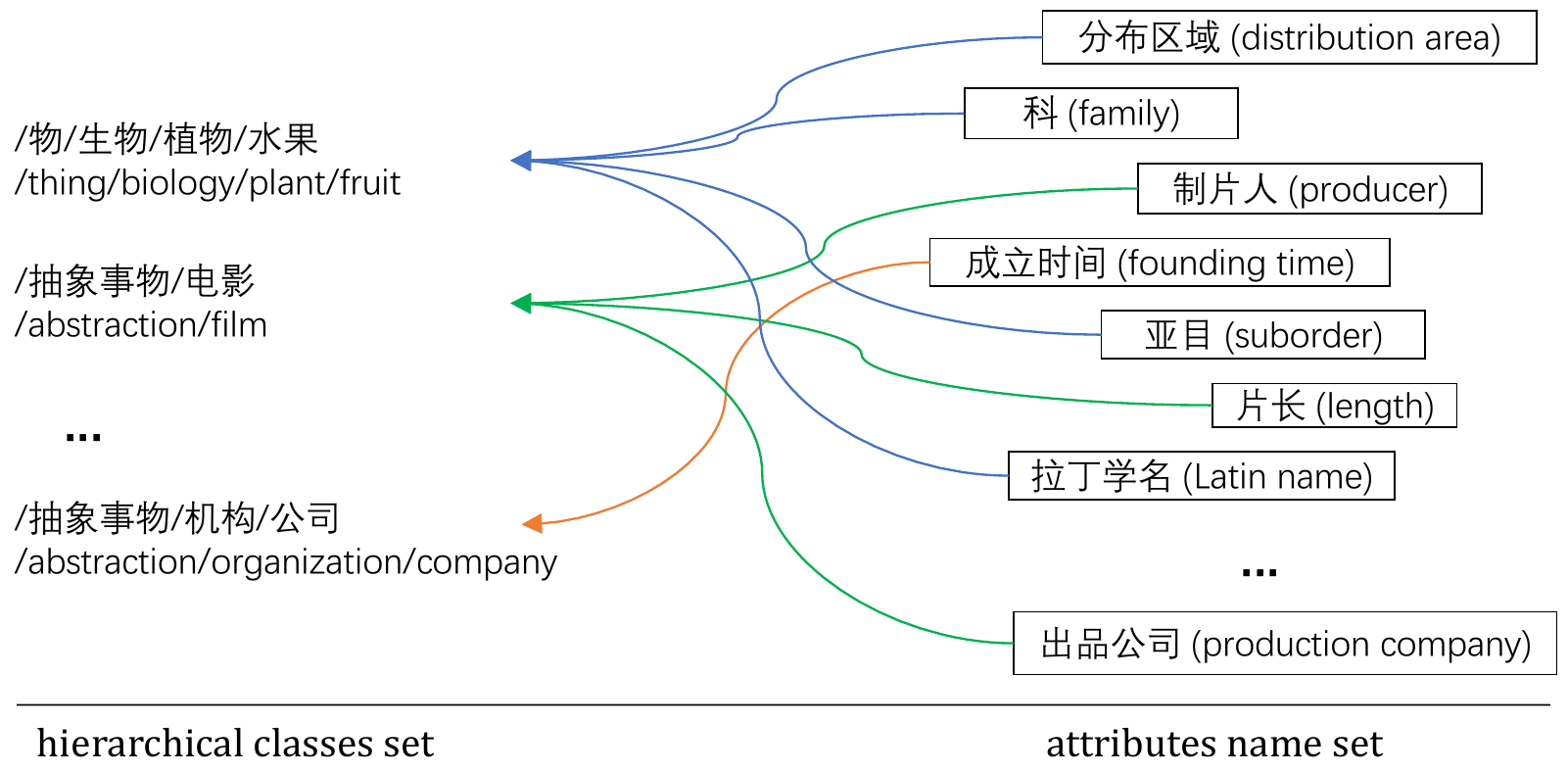}
	\end{center}
	\caption{Attribute Acquisition for Hierarchical Classes.}
	\label{fig: task}
\end{figure}

For Artificial Intelligence and Semantic Web researchers, 
an ontology is a document or database that formally defines 
the characteristic of terms and relations among them for knowledge base
~\cite{berners2001semantic,shadbolt2006semantic,lehmann2015dbpedia}, 
like the schema for relational database. \textit{Attribute} 
plays a crucial role in ontology, describing the characteristic 
of given \textit{class} and connects different \textit{entities} 
in knowledge base (for example, ``\texttt{director}'' is an 
attribute for class ``\texttt{film}'', which also connects 
entities of class ``\texttt{film}'' and ``\texttt{person}''), 
here \textit{class} means the word describing type of an entity.
Thus attribute acquisition is essential for the task of 
ontology construction, acquiring attributes for the given 
hierarchical classes formed by hypernym-hyponym relation  
(i.e., \textit{is-a} relation). The task shown in Fig.~\ref{fig: task}. 
Here, we think the entity can be regarded as instance of 
its classes, thus the attributes of an entity can be acquired 
by inspecting its classes. For an enclosed or specific 
domain knowledge base, experts could acquire the attributes 
by combing the guidance of the meaning of classes in ontology 
and their expert-level background knowledge, however, such 
approach is label intensive and time-consuming, which is 
not scalable for an open domain knowledge base, due to 
the continuous update of domain categories and entity classes.

An effective and simple solution for attribute acquisition 
in open domain knowledge base is community-collaborated method. 
The typical examples are DBpedia
~\cite{lehmann2015dbpedia}, FreeBase~\cite{bollacker2008freebase} 
and Wikidata~\cite{vrandevcic2014wikidata} etc. However, for Chinese 
knowledge bases, there is no such huge community to build 
and maintain the ontology within short time. Although 
various data-driven methods, deriving a large number of 
facts from large-scale corpora by automatic extraction, 
have been utilized to build large-scale open domain Chinese 
language knowledge bases~\cite{niu2011zhishi,wang2013xlore,xu2017cn-pedia},
most of them reuse the existing taxonomy of online encyclopedias 
(such as Baidubaike\footnote{https://baike.baidu.com/, 
one of the largest Chinese language encyclopedias.}) 
to build ontology, which is static and usually coverage-limited.

In this paper, We attempt to propose an automatic paradigm 
to acquire attributes depending on hierarchical classes, 
further for significantly improve the scalability of 
ontology and efficiency of building a knowledge base.
Typically, compared with the meaning expressed by only one 
word to represent one class (e.g. ``\texttt{university}'' to indicate 
this class is a school for high education), the meaning of 
one \textit{class-path} (several words connected through a path, 
e.g. ``\texttt{institute/school/university}'', the former one is the 
hypernym of the latter one) is more explicit and solid. 
For example, ``\texttt{university}'' also means the body of faculty 
and students at a university, with the help of ``\texttt{institute}'' 
and ``\texttt{school}'', the meaning of university is refined. For 
this reason, we utilize class-path to represent the hierarchical 
classes in ontology, instead of the terminal class word
of the hypernym-hyponym relation (i.e., \textit{is-a} relation) 
based hierarchy. Inspired by recent representation learning 
of knowledge base
~\cite{bordes:2013,wang2014TransH,lin2015TransR,xiao2015Manifold}, 
we consider mapping the hierarchical classes and attributes 
into continuous vector space, and transforming the attribute 
acquisition problem into a prediction task.

Traditional data-driven method for knowledge base construction 
mostly allocate attributes to entities, which is damaged by 
the inconvenience so called multi-role entities. (e.g. 
there are two classes for every entity on average 
in BigCilin \footnote{http://www.bigcilin.com/. BigCilin 
contains over 9 millions entities and about 12 millions 
hypernym-hyponym relation (i.e., ``is-a'' relation) pairs
with over 45 thousands classes concept.} , a Chinese open 
domain hypernym-hyponym based knowledge base, with nearly 
4.5 million entities having more than two classes). 
For example, the entity ``\texttt{apple}'' can refer to 
both fruit and company, or even a movie, actually, there 
are about 12 meanings for the entity ``\texttt{apple}'' 
in Baidubaike. Some entities referring to the same things 
might also have different roles, such as the entity 
``\texttt{Ronald Wilson Reagan}'', can indicate both 
actor and president of American. In fact, unlike the 
value of attribute, e.g. the concrete ``\texttt{color}'', 
like ``\texttt{green}'', is the value of the attribute 
``\texttt{color}'' of the entity ``\texttt{apple}'', 
the attribute is more general. For the entities 
in the same class, they almost share the same attribute 
sets. For example, ``\texttt{apple}'' and ``\texttt{pear}'' 
under the class ``\texttt{fruit}'' have the same attribute 
sets. That means different from allocating attribute to 
entities, allocating attribute to class seems more sense.

\textit{Ambiguous attributes-allocation problem} caused 
by multi-role entities between classes and attributes 
is the major challenge 
of our work. For instance, there is an attribute of 
``\texttt{director}'' for entity ``\texttt{apple}'', but 
the actual class for the attribute is unknown,  due to 
the multiple roles of ``\texttt{apple}', such as 
``\texttt{fruit}'', ``\texttt{film}'', or 
``\texttt{company}'', etc. Directly obtaining the 
corresponding pair of classes and attribute will lead to 
much noises during the training process, for example, 
``\texttt{director}'' is not an attribute for ``\texttt{fruit}'', 
even they are bridged by entity ``\texttt{apple}''. 
In this paper, we propose a translation-based embedding 
method based on selective attention model, called 
\textbf{TransAtt}, with the ability of representation
learning of hierarchical classes and attributes. 
Though our work is focused on Chinese, the proposed 
method is language-independent for attribute acquisition 
in ontology of knowledge base.

The main contributions of this paper are as follows:

\textit{\romannumeral1)} Constructing a dataset called 
\textbf{BigCilin11K}, to the best of our knowledge, this 
is the first dataset with abundant hierarchical classes 
and attributes of entities in Chinese language.

\textit{\romannumeral2)} Presenting an automatic paradigm 
for attribute acquisition by learning the continuous 
representation of hierarchical classes and attributes, 
called \textbf{TransAtt}, which certainly relieves the 
burden of manual attribute acquisition.

\section{Related Work}


\subsection{Ontology Construction}

For an open domain knowledge base, the ontology cannot be modeled once for all, because it changes when new facts populated. 
The light-weight knowledge base YAGO is constructed based on WordNet~\cite{miller1995wordnet} and Wikipedia~\cite{suchanek2007yago}, 
but it is limited in coverage of ontology when new facts emerge.
Many other popular knowledge bases, like DBpedia~\cite{lehmann2015dbpedia}, 
FreeBase~\cite{bollacker2008freebase} and Wikidata~\cite{vrandevcic2014wikidata} etc., 
adapt community-collaborated method to maintain and extend the ontology on a website.
As time goes on, a mass of volunteers have been accumulated for English knowledge base. 
However, for Chinese knowledge bases there is no such huge community to 
construct such a satisfactory ontology.

In recent year, there are many Chinese knowledge base being developed,
but there still a huge research gap in the construction of ontology.
Some famous Chinese language knowledge bases, 
like zhishi.me~\cite{niu2011zhishi}, XLore~\cite{wang2013xlore}, and CN-pedia~\cite{xu2017cn-pedia},
focus on the extraction of entity-relation triples without constructing powerful and dynamic ontology. 
They utilize existing category information provided in encyclopedias (e.g., Wikipedia, Baidubaike) to construct ontology,
which will be failed when new classes injected. BigCilin is a Chinese open domain hypernym-hyponym based knowledge base 
developed by Harbin Institute of Technology, which will dynamically construct its architecture of hierarchical classes when new entities are injected.
However BigCilin does not have a complete ontology even with a hierarchical classes architecture, such as attribute-allocation of the classes. 


\subsection{Knowledge Base Representation Learning}

Knowledge base representation learning aims at offering a continuous knowledge representation paradigm by
transforming the entities and relations into continuous vector space. 
\textit{Translation-based embedding methods} are adopted extensively in representation learning of 
knowledge base~\cite{bordes:2013,wang2014TransH,lin2015TransR,xiao2015Manifold},
sharing the core translation principle $\mathbf{h+r=t}$~\cite{bordes:2013} for the triple $(h,r,t)$, where $h,r,t$ indicate
a head entity, a relation, and a tail entity, respectively, and their embedding vectors are $\mathbf{h,r,t}$.
The basic idea behind the principle is that the relation $r$ corresponds to a translation of the embeddings, translating
the head entity to a tail entity. Unlike the entity-relation triple, 
class-attribute tuple has no explicit translation operator.
So we just induce an attribute-corresponding matrix as translation operator, mapping given class to attributes space.

Besides, getting the corresponding pairs of class and attribute for training is unobtainable, 
due to the ambiguous attribute-allocation problem between class and attribute in data-driven knowledge bases. 
For this issue, a representation learning method with the ability of selection over classes should be considered, 
as this paper to be discussed.

\section{Preliminaries}

In this section, firstly, we introduce some concepts 
occurred in this paper and a brief overview of the 
notation representation of the subset of knowledge 
base. Then, formal description of our task will be 
presented.


\textbf{Ontology}  We refer ontology as an architecture 
of knowledge base defining the characteristic of 
terms and relations among them. like the schema for 
relational database.

\textbf{Class}  A class is the word describing the type 
of an entity, for example, ``\texttt{fruit}'' is a class of the 
entity ``\texttt{apple}', and they form a hypernym-hyponym 
relation.

\textbf{Class-Path}  Representing the hierarchical 
classes of hypernym-hyponym based relations.  
The nodes of the path are classes, and the edges 
represent hypernym-hyponym relations between 
the nodes.

We investigate a subset of knowledge base 
involving classes and attributes information of entities, without the relational facts of entities. 
Formally, the subset denoted as $\mathcal{K}$, 
and $\mathcal{K}=\{\mathcal{E,C,A}, \mathcal{R}_1, \mathcal{R}_2, \mathcal{R}_3\}$,
where $\mathcal{E}$, $\mathcal{C}$ and $\mathcal{A}$ are the set of entities, classes and attribute name respectively. 
$\mathcal{R}_1$ is the set of hypernym-hyponym relation in ontology with two elements 
denoted as $(c_k, e_i)$ and $(c_j, c_i)$ respectively, 
where $e_i \in \mathcal{E}, c_i \in \mathcal{C}$, and their classes $c_k, c_j \in \mathcal{C}$.
$\mathcal{R}_2$ is the attribute-allocation for entities
where each tuple element is denoted as $(e_i, a_j)$, where $e_i \in \mathcal{E}$, 
and $a_j \in \mathcal{A}$, which means that the entity $e_i$ has the attribute $a_j$.

\begin{figure}[htb]
	\begin{center}
		\includegraphics[width=14cm]{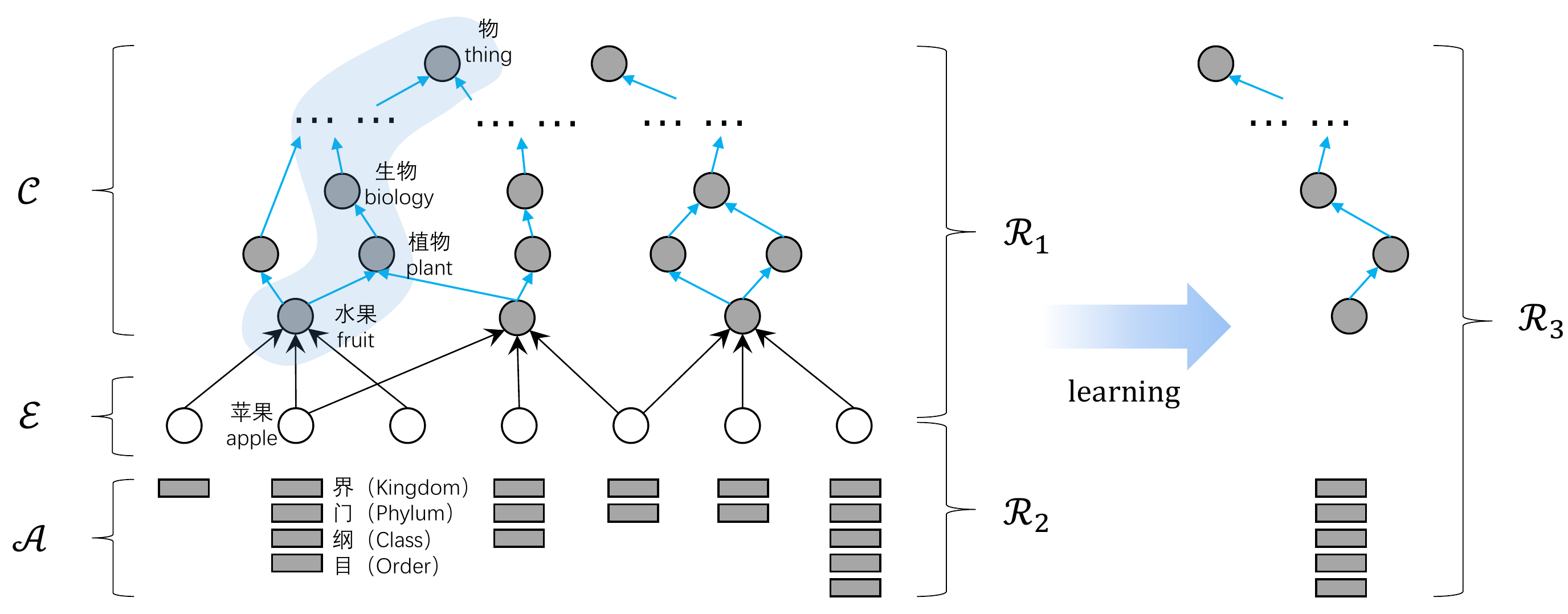}
	\end{center}
	\caption{Formal Task Definition with Notation Representation of $\mathcal{K}$.}
	\label{fig: KB1}
\end{figure}

Given $\mathcal{E,C}, \mathcal{R}_1$, we can get some class-paths for $\forall e\in \mathcal{E}$, 
the $k$-th class-path of $e$ is represented by $p_k^e$, and $p_k^e = (c_1^e, c_2^e,\cdots,c_i^e, c_{i+1}^e,\cdots,c_n^e)$,
where $n$ is the length of class-path $p_k^e$, also the number of classes in the path. 
For $i\in [1,n-1], c_i^e, c_n^e \in \mathcal{C}$ and $(c_i^e, c_{i+1}^e), (c_n^e, e)\in \mathcal{R}_1$.
One of class-paths is labeled with shadow in Fig.~\ref{fig: KB1}.
In the remainder part, we use class-paths to represent the hierarchical classes.
And $\mathcal{R}_3$ is the attribute-allocation for class-paths 
with each tuple element denoted as $(p_i, a_j)$, where $p_i$ is one of class-path in $\mathcal{K}$ and
$a_j \in \mathcal{A}$, meaning the class-path $p_i$ has the attribute $a_j$. 
The left part of arrow in Fig.~\ref{fig: KB1} illustrates the notation representation of $\mathcal{K}$.



\textbf{Task definition}  Formally, given $\{\mathcal{E,C,A}, \mathcal{R}_1, \mathcal{R}_2\}$ of $\mathcal{K}$ from knowledge base, 
we are supposed to predict $\mathcal{R}_3$, and further use $\mathcal{R}_3$ for $\mathcal{R}_2$ completion.
The prediction of $\mathcal{R}_3$ can be regarded as attributes prediction for class-paths, 
and the completion of $\mathcal{R}_2$ based on $\mathcal{R}_3$  
is attributes prediction of entities, which is achieved by inspecting their classes. This procedure is illustrated in Fig.~\ref{fig: KB1}.

\section{Methodology}

The main idea is to apply a sequence model to the class-path, 
then jointly learn both representation of attributes and class-paths 
based on a translation-based embedding model together with selective attention over the class-paths,
called \textbf{TransAtt}, to deal with \textit{ambiguous attributes-allocation problem} between class-paths and attributes.
 In the end of this section, the training algorithm of TransAtt will be presented.


\subsection{Representation Learning of Class-Path via LSTM}\label{model-class-path}

Since a class-path is a sequence 
of class words, a sequence model could be considered.
Recurrent Neural Network (RNN)~\cite{Mikolov:2010} is a type of artificial neural network 
whose connections between neurons form a directed cycle, 
and create a hidden state of the network which allows it to exhibit dynamic temporal 
behavior according to the history information. This feature makes RNN suitable for sequential data 
such as unsegmented connected handwriting or speech recognition. 
The most representative variant of RNN is Long Short Term Memory (LSTM)~\cite{Hochreiter:1997} 
which is capable of learning long-term dependencies. 

\begin{figure}[htb]
	\centering 
	\includegraphics[width=9cm]{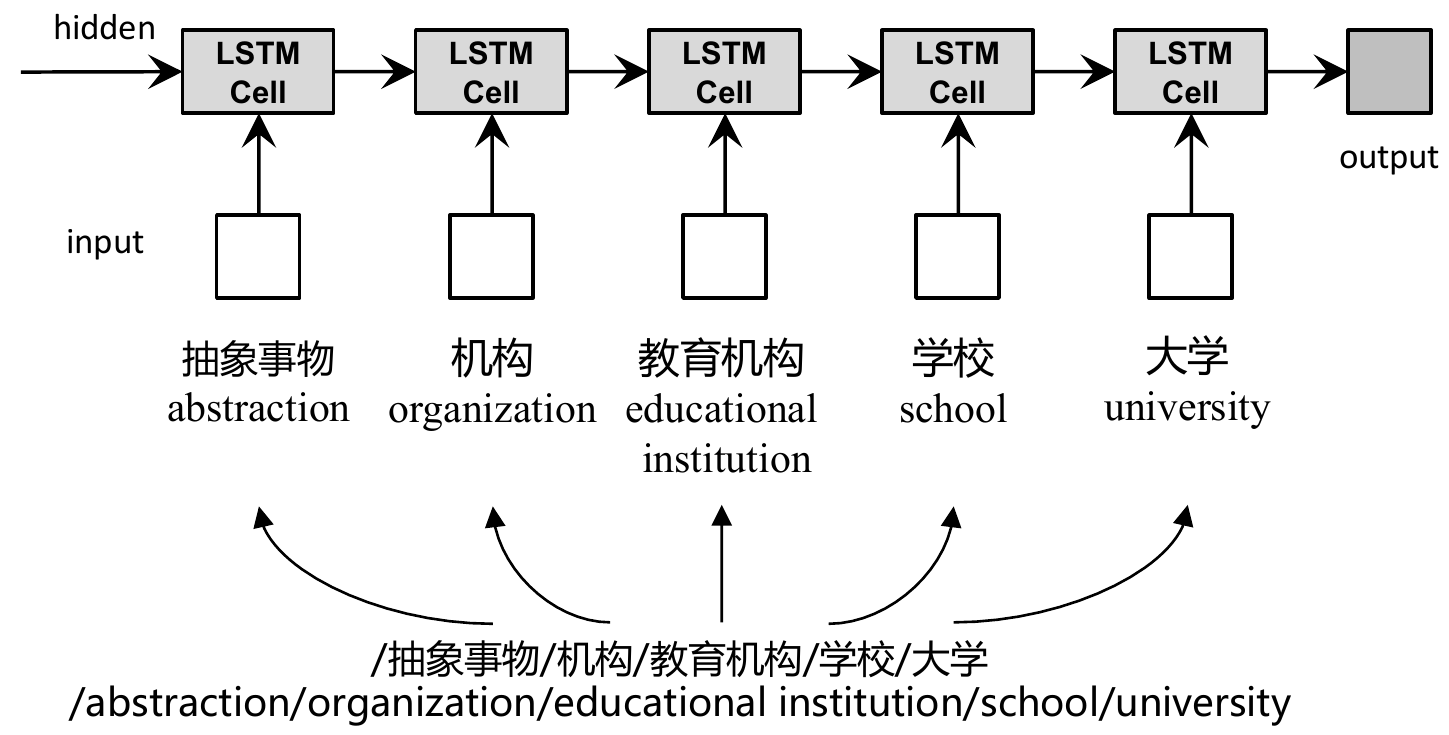}
	\caption{A LSTM network for Class-Path Representation.}
	\label{fig: LSTM-class-path}
\end{figure}

In this paper, we use LSTM to learn the representation of class-path. LSTM is nearly identical as RNN,
except that the updates in hidden layer are replaced by different types of memory cells to store information, 
which makes it better at exploiting long term context.  
The LSTM-cell is implemented as Graves et al.~\shortcite{graves:2005}. 
Fig.~\ref{fig: LSTM-class-path} shows a LSTM sequence model which employs aforementioned LSTM-cell (light gray boxes),
applied by the class-path \texttt{"/抽象事物/机构/教育机构/学校/大学"}, which means 
\texttt{"/abstraction/organization/educational institution/school/university"} in English. 
The input is pre-Strained embedding vectors~\cite{Mikolov:2013a} of the class words in class-path,
and the final output (dark gray boxes) of the sequence model is the vector representation of a given class-path.

\subsection{TransAtt: a Translation-Based Embedding via Selective Attention Model}\label{model-TransAtt}

Unlike conventional translation-based embedding methods which focus on entity-relation triples $(h,r,t)$, 
the focus in our work is the two-elements tuples $(p,a)$ formed by different class-path and its attribute. 
Therefore, there is no explicit translation operator like $r$ in triple $(h,r,t)$. 
To solve this problem, we construct a mapping-matrix for each attribute, and the 
translation process is $\mathbf{p}\mathbf{M}_a = \mathbf{a}$, where $\mathbf{p, a},\mathbf{M}_a$ 
are the embedding vectors of class-path, attribute and its mapping-matrix respectively.

\begin{figure}[htb]
	\centering 
	\includegraphics[width=16.5cm]{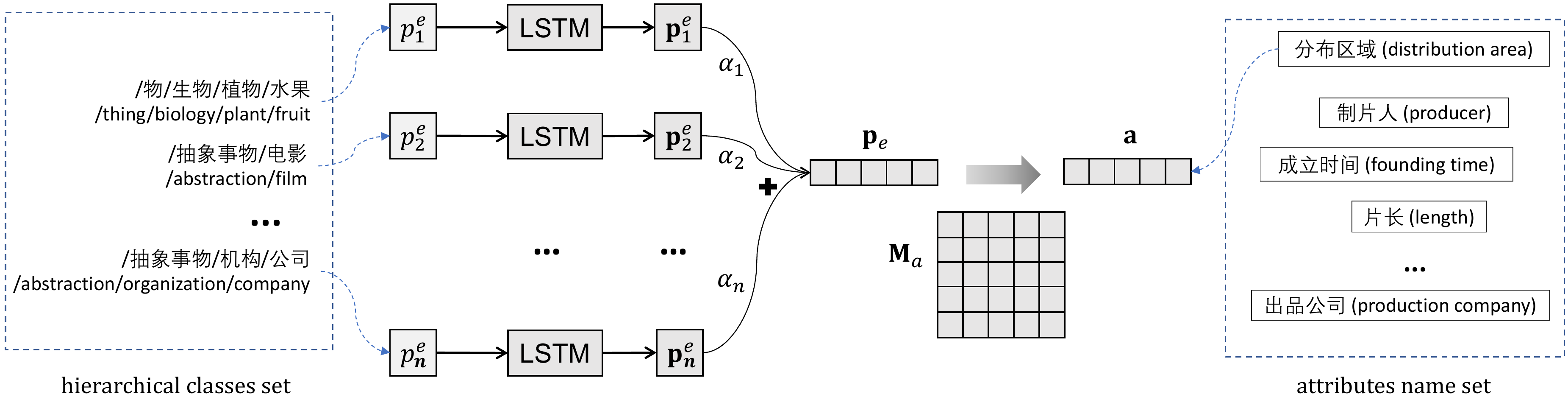}
	\caption{Translation-Based Embedding via Selective Attention Model.}
	\label{fig: TransAtt}
\end{figure}

From $\mathcal{K}$, we can get a set of tuples in form of $(P_e, a)$, where $P_e$ is the set of class-paths for a given entity $e$ and 
$a$ is an attribute the entity. However, we do not know which of these class-paths in the $P_e$ has attribute $a$.
In order to deal with the ambiguous problem between class-path and attribute as well as to learn their representation jointly, 
we propose an attention-based translation embedding model, called \textbf{TransAtt},
which is composed of three parts: (\romannumeral1) representation learning model for class-path; 
(\romannumeral2) selective attention model over class-paths; 
and (\romannumeral3) translation-based embedding model to predict attributes.
An overview of TransAtt is shown in Fig.~\ref{fig: TransAtt}.
Part (\romannumeral1) have been introduced in Sec.~\ref{model-class-path}, 
and we now proceed to some details for the last two parts.

\textbf{Selective attention model over class-paths } 
As we have mentioned, $P_e$ is the set of class-path for a given entity $e$, i.e., $P_e=\{p_1^e,p_2^e,...,p_i^e,...,p_n^e\}$, 
where $p_i^e$ is one of class-paths of $e$, and there are a total of $n$ class-paths of $e$. 
To exploit the information of all class-paths, our model represents the set $P_e$ 
with a real-valued vector $\mathbf{p}_e$ during the prediction of attribute $a$. 
$\mathbf{p}_e$ depends on the representations of all class-paths 
$\mathbf{p}_1^e,\mathbf{p}_2^e,...,\mathbf{p}_i^e,...,\mathbf{p}_n^e$,
thus it can be computed as a weighted sum 
of these representations by $\mathbf{p}_e = \sum_{i}\alpha_{i}\mathbf{p}_i^e$,
where $\alpha_i$ is the weight of each representation $\mathbf{p}_i^e$. 
Due to the ambiguous attributes-allocation between class-path and attribute, we cannot regard each class-path equally
because the inaccurate attributes-allocation will lead to noises during training. 
Hence, we use the selective attention to de-emphasize the noisy class-path, and set
$\alpha_{i} = \frac{\mathrm{exp}(s_i)}{\sum_{k}\mathrm{exp}(s_k)}$, 
where $s_i$ is used to score how well the input class-path $p_i^e$ and
the predict attribute $a$ matches. 
We select the bilinear form which achieves the best performance among different alternatives:
$s_i = \mathbf{p}_i^e\mathbf{Aa}$, where $\mathbf{A}$ is a learnable weighted matrix,
and $\mathbf{a}$ is the embedding vector of attribute $a$.

\textbf{Translation-based embedding model to predict attributes } 
For the given attention-based representation $\mathbf{p}_e$ the set of class-paths, 
the translation-based embedding model first utilizes the simple linear transformation to 
map the attention-based representation $\mathbf{p}_e$ into attribute vector space, 
then predict the attribute name based on some distance measure in the vector space.
Now we can rewrite the previous simple translation process 
as $\mathbf{p}_e\mathbf{M}_a = \mathbf{a}$, where $\mathbf{p}$ is replaced by $\mathbf{p}_e$
and $\mathbf{M}_a$ is the attribute-corresponding mapping-matrix. 

\subsection{Training Algorithm}

According to the translation-based embedding model, 
we want $\mathbf{p}\mathbf{M}_a = \mathbf{a}$ when $(p,a)$ holds (i.e., $\mathbf{a}$ should be 
one of the nearest neighbors of $\mathbf{p}\mathbf{M}_a$), 
otherwise $\mathbf{p}\mathbf{M}_a$ should be far away from $\mathbf{a}$. 
To achieve that, we need an energy-based framework, the energy of a tuple is equal to 
$d(\mathbf{p}\mathbf{M}_a, \mathbf{a})$ for some dissimilarity measure $d$, 
which we set to either L1 or L2-norm.

Given $\mathcal{K}$, we can collect a set of tuples $(P_e, a)$ as training set $\Omega$,
where $P_e$ is the set of class-path for a given entity $e$ and 
$a$ is an attribute of $e$. 
We then minimize a margin-based ranking criteria over the training set 
to jointly learn the representation of class-paths and attributes:

$$\mathcal{L}=\sum_{(P_e, a)\in \Omega}\sum_{(P_e, {a}')\in \Omega'_{(P_e, a)}} [\gamma +d(\mathbf{p}_e\mathbf{M}_a, \mathbf{a})-d(\mathbf{p}_e\mathbf{M}_{a'}, \mathbf{a'})]_+$$

\noindent where  $[x]_+$ denotes the positive part of $x$, $\gamma > 0$ is a margin hyperparameter, $a'$ is the corrupted attribute for $P_e$, and 
the set of corrupted tuples $\Omega'_{(P_e, a)}=\{ (P_e, a') | a'\in \mathcal{A} \}$, is composed of training tuples
with attribute replaced by a corrupted attribute, which is selected from $\mathcal{A}$ randomly.

To train the neural networks in TransAtt, we utilize backpropagation~\cite{Rumelhart:1985} to compute the gradient of parameters. 
We then apply Adadelta~\cite{Zeiler:2012} to perform optimization. 
Adadelta is an extension of Adagrad~\cite{Dhillon:2011} that aims to reduce its aggressive, monotonically decreasing learning rate. 
Instead of accumulating all past squared gradients, Adadelta restricts the window of accumulated past gradients to a certain fixed size. 

\section{Experiments}

In the experimental stage, we implement and train the TransAtt to learn the embedding representations of class-paths and attributes. 
We first outline the preparation in experimental setup, 
including a dataset called \textbf{BigCilin11K} which we construct manually, as well as the evaluation metrics.
We empirically study and evaluate TransAtt on two tasks: attributes prediction for entity (abbr. \textbf{APE}), 
and attributes prediction for class-path (abbr. \textbf{APC}).
These two tasks evaluate the performances of our representative method from two diverse directions. 
In detail, APE focuses on evaluating TransAtt's capacity of selection over classes for attributes of multi-roles entities, 
while APC  focuses on the precision of attributes prediction for class-path based on TransAtt's representation learning.
Finally, we report the performance and perform visualization on our model.

\subsection{Dataset Construction}

BigCilin has abundant information of hierarchical classes and the capacity of dynamically 
building the hierarchical structure given unknown entities.
Furthermore, BigCilin is a data-driven knowledge base which extracts attribute facts from encyclopedias (like info-box of Baidubaike). 
Based on BigCilin, we construct a dataset called \textbf{BigCilin11K} as follow:

(1) Randomly extract 20,000 entities and their classes. Based on these classes, obtain the entire database of hypernym-hyponym relations in the class-level from BigCilin.
i.e., $\mathcal{E,C}, \mathcal{R}_1$ of $\mathcal{K}$. Then, filter the entities without attribute and approximately 11,000 entities are left.
i.e., $\mathcal{A}, \mathcal{R}_2$ of $\mathcal{K}$.

(2) Construct and store the class-paths of every entity 
based on the extracted hypernym-hyponym relation set $\mathcal{R}_1$ at entity-level and class-level.

(3) Filter the attributes with low frequency. Select the attributes which have appeared in at least 20 entities 
(the threshold is determined by the coverage of entity with attributes).

We also randomly select 3,000 entities to evaluate the task of attributes prediction for entities 
and construct a class-path set including 240 class-paths that are unseen in training stage. 
The statistics of the dataset BigCilin11K is shown in Tab.\ref{tab: statistics}, 
where APE and APC represent the tasks of attributes prediction for entities and class-path respectively.

\begin{table}[htb]
	\setlength{\abovecaptionskip}{0.cm}
	\setlength{\belowcaptionskip}{-0.cm}
	\begin{center}
		\renewcommand{\multirowsetup}{\centering}
		\renewcommand{\arraystretch}{1.2}
		\setlength\tabcolsep{4pt} 
		\begin{tabular}{|c|c|c|c|}
			\hline
			- & \bf \# Class-Paths & \bf \# Entities & \bf \# attributes \\
			\hline 
			\bf Trian & 5,174 & 11,161 & \multirow{2}{1.5cm}{286} \\
			\bf Test & 240 (for APC) & 3,000 (for APE) & \\
			\hline
		\end{tabular}
	\end{center}
	\caption{\label{tab: statistics} The Statistics of BigCilin11K. }
\end{table}

\subsection{Evaluation Metrics}

We adopt evaluation criteria in information retrieval field,
because the tasks of attributes prediction from a candidate attribute set are considered as retrieval tasks.
In this paper, \textit{P@k} and \textit{Hits@k} are utilized to evaluate performance in the two tasks.

\textit{P@k} measures the precision of top-$k$ returned results, 
corresponding to the number of relevant attributes in top-$k$ returned results.
e.g., \textit{P@10} calculates the proportion of relevant attributes in top-10 returned results, 
if there are 7 relevant attributes, \textit{P@10} equals to $7/10=0.7$.
In order to measure the overall performance on \textit{P@k}, 
we calculate the mean value as ${\sum_{n=1}^{Q}P@k(q)}/{Q}$, 
where $Q$ is the number of entities or class-path.

\textit{Hits@k} measures the proportion of entities attaining at least one relevant attribute in the top-$k$ returned results. 
e.g. if there are 100 entities and 88 of them attain at least one relevant attribute 
in the top-10 returned results, then the \textit{Hits@10} is $88/100=0.88$.

\subsection{Attributes Prediction for Entity (APE)}

Viewing from an entity, predicting attributes for it is undoubtedly necessary and important. 
A more exciting work is bonding the predicted attributes with some class-paths of the entity,
because the entity might have multi-roles, like the example of ``\texttt{苹果(apple)}'' mentioned before.
Therefore, we propose the task of attributes prediction for entities (abbr. \textbf{APE}), 
in order to predict the attributes for given entity as well as bond them with relative class-paths. 

Since the entity is regarded as instance of its classes,
the set of attributes for an entity can be obtained by merely inspecting its classes,
especially for fine-grained classes. In APE, for a given entity, we
first obtain its class-paths from $\mathcal{K}$, then use them to predict the attributes of entity.
The prediction phase utilizes the entire TransAtt model with parameters and representation learned.
\textit{Hits@k} evaluation method is adapted for APE
(filtering of common attributes, like ``\texttt{中文名(Chinese name)}'', ``\texttt{外文名(English name)}'', etc.)
because there is no complete standard entity-attribute dataset 
but only some incomplete entities' information extracted from 
Baidubaike's info-box for the 3,000 test entities (Tab.~\ref{tab: statistics}) in the evaluation stage.

Besides inspecting \textit{Hits@k} on overall entities, 
different categories of entities are also considered 
(i.e., ``\texttt{Things}'', ``\texttt{Abstraction}'', ``\texttt{Person}''). 
Final results are shown in Tab.~\ref{tab: Hits}, 
where number in brackets are the number of entities of corresponding category.
Results indicate that nearly 76\% of overall entities have obtained 
at least one info-box attribute in top-20 predicted attributes. 
For entities of ``\texttt{Person}'' type, the results of \textit{Hits@k} are all 
over 80\% except \textit{Hits@1}. For the other two categories, 
the performance is poor as compared to overall evaluation.
The results in Tab.~\ref{tab: Hits} is evaluated under the worse situation, 
due to the incompleteness of info-box producing such as false negative noise. 
A more suitable method \textit{P@k} will be manually evaluated 
in task APC, but it is not automated and hence time-consuming.

\begin{figure*}[htb]
	\begin{center}
		\includegraphics[width=16cm]{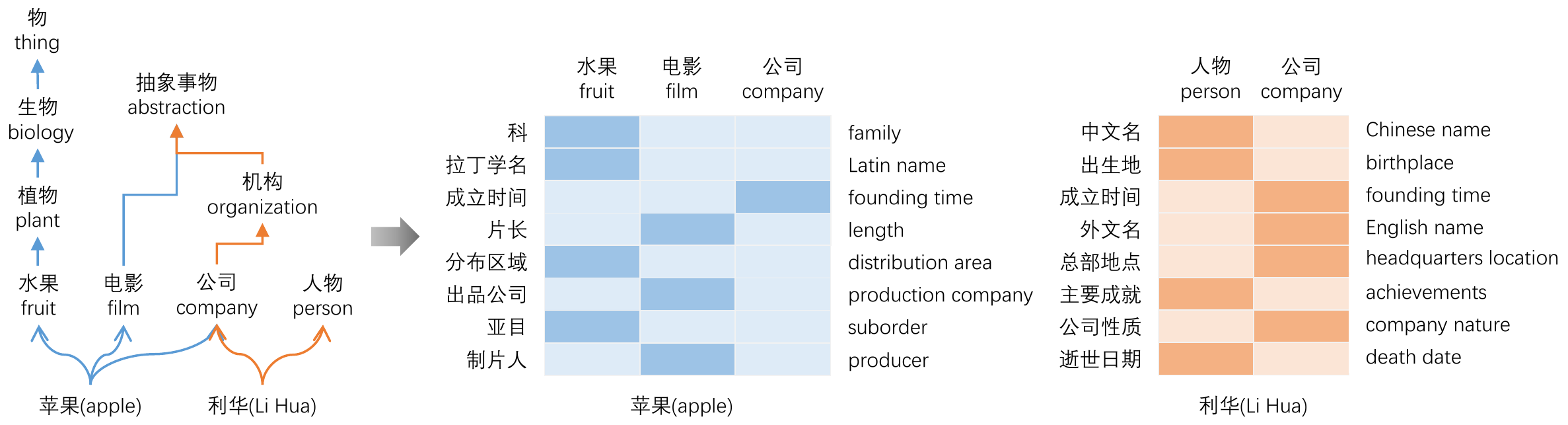}
	\end{center}
	\caption{Visualization of top Predicted attributes of Multi-Roles Entities.}
	\label{fig: entity_cases1}
\end{figure*}

Since we are more concerned about multi-roles entities in APE, which requires bonding predicted attributes with relative class-paths, 
we further illustrate the predicted attributes by our model for two multi-roles entity samples in Fig.~\ref{fig: entity_cases1},
namely, ``\texttt{苹果(apple)}'' and ``\texttt{利华(Li Hua)}''.
Class-paths for the two entities revealing the multi-roles nature are illustrated on the left hand side (e.g., tree class-paths for entity ``\texttt{苹果(apple)}'' 
representing roles of fruit, film, and company respectively).
Following the arrow, we outline the top predicted attributes 
and two matrices representing the selective class-paths
\footnote{for economizing space, here we use class name to represents related class-path in attention matrix part in Fig.~\ref{fig: entity_cases1}.} 
attention weights of entities. 
Different columns of matrix represents the attention of predicted attributes for different class-paths, where darker cell indicates higher weights. 
From the attention matrices, it is observed that the predicted attributes attend the relative class-paths and nearly zero on uncorrelated ones, 
e.g., an accurate example is the class-path ``\texttt{/abstraction/film}'', which gets the attentions of attributes ``\texttt{length, production company, producer}''.
However, there are also some mistakes: the attributes ``\texttt{Chinese name}'' and ``\texttt{English name}'' only attend one of the class ``\texttt{person}'' or ``\texttt{company}'', which is inaccurate because both of them should have these two attributes in common sense.
Fortunately, such mistake can be remedied by the task of APC, which is from viewpoint of class to execute the prediction.

\begin{table}[htbp]
	\begin{center}
		\begin{minipage}[t]{.45\linewidth}  
			\centering
			\renewcommand{\arraystretch}{1.2}
			\setlength\tabcolsep{3pt}
			\begin{tabular}{|lccc|c|}
				\hline
				\multirow{2}{0.5cm}{\bf Hits@k} & \bf Thing & \bf Abstr. & \bf Person & \bf Overall \\
				&(848)&(1,505)&(727)&(2,641) \\
				\hline
				\bf Hits@1 & 26.30 & 26.51 & 64.37 & 36.80 \\
				\bf Hits@5 & 50.71 & 46.18 & 80.47 & 56.57 \\
				\bf Hits@10 & 60.73 & 56.35 & 84.73 & 65.05 \\
				\bf Hits@15 & 67.92 & 63.85 & 87.48 & 71.34 \\
				\bf Hits@20 & 73.58 & 73.58 & 89.27 & 75.99 \\
				\hline
			\end{tabular}
		\caption{\label{tab: Hits}\textit{Hits@k} Evaluation on APE task. }
		\end{minipage}
		\begin{minipage}[t]{.5\linewidth}
			\centering
			\renewcommand{\arraystretch}{1.2}
			\setlength\tabcolsep{3pt}
			\begin{tabular}{|lccc|c|}
				\hline
				\multirow{2}{1.8cm}{\bf mean P@k} & \bf Thing & \bf Abstr. & \bf Person & \bf Overall \\
				&(59)&(71)&(15)&(184) \\
				\hline
				\bf mean P@1 & 77.97 & 83.10 & 73.33 & 78.26 \\
				\bf mean P@5 & 70.85 & 77.46 & 68.00 & 73.37 \\
				\bf mean P@10 & 65.08 & 68.73 & 72.67 & 67.45 \\
				\bf mean P@15 & 60.34 & 62.63 & 68.89 & 62.43 \\
				\bf mean P@20 & 53.64 & 55.92 & 61.67 & 55.76 \\
				\hline
			\end{tabular}
		\caption{\label{tab: Precision}\textit{P@k} Evaluation on APC task. }
		\end{minipage}
	\end{center}
\end{table}

\subsection{Attributes Prediction for Class-Path (APC)}

In the viewpoint of class-path, we propose the task of attributes prediction for class-path (abbr. \textbf{APC}) in order 
to evaluate the capacity of model for mapping class-paths to attributes, 
which is helpful for dynamic ontology construction. 
During the prediction process of this task, we utilize TransAtt model without selective attention model over class-paths,
as each prediction only involves a unique class-path and hence selective attention is not needed.

Currently, there is lack of support for standard complete class-attribute pairs dataset in Chinese language, 
so we manually label ``\texttt{True}'' or ``\texttt{False}'' for top-$k$ predicted attributes
(excluding common attributes such as ``\texttt{中文名(Chinese name)}'', ``\texttt{外文名(English name)}'').
The manual labeling principles are follows:

(1) Filter inaccurate class-paths that include fake class word or are not consistent with the hierarchy of hypernym-hyponym.

(2) Filter class-path that are used to describe an abstract concept but not concrete entity,
e.g., ``\texttt{/抽象事物/能力/竞争力(/abstraction/capacity/competitiveness)}''.

(3) Utilize search engine if meet with any unfamiliar domain or word.

After filtering, 184 of 240 test class-paths (see Tab.~\ref{tab: statistics}) are included in \textit{P@k} evaluation.
Final results are shown in Tab.~\ref{tab: Precision}, where number in brackets are the number of entities of corresponding category after filtering.
From the results, it is observed that the continuous representation learned can be utilized to predict attributes accurately for class,
with higher than 75\% of \textit{P@1} in overall class-paths and nearly 70\% of \textit{P@10} of different categories. 
These given class-paths are unseen in training stage, but can be mapped to relatively accurate set of attributes,
which indicates that the construction of ontology can be dynamic.
We also outline some examples from prediction results in Tab.~\ref{tab: class2attribute} with Chinese and English language labeled. 
Note that they are in different domains, but our model can predict domain relative attributes for them accurately.

\begin{table*}[htb] \scriptsize
	\begin{center}
		\renewcommand{\arraystretch}{1.4}
		\setlength\tabcolsep{3pt}
		\begin{tabular}{|l|}
			\hline
			\textbf{Class-Path: /抽象事物/戏剧/歌曲/名曲(/abstraction/drama/songs/famous songs) }\\ 
			\textbf{top-10 attributes:}发行时间(issue date), 歌曲时长(duration), 编曲(arranger), 歌曲语言(language), 填词(lyricist),\\
			所属专辑(album), 歌曲原唱(original singer), 谱曲(composer), 音乐风格(style), 出生日期(birth date)\\
			\hline
			\textbf{Class-Path: /抽象事物/组织/医院/烟台市医院(/abstraction/organization/hospital/hospital of Yantai)} \\ 
			\textbf{top-10 attributes: }地理位置(location), 成立时间(founding time), 所属地区(affiliating area), 地址(address), 医院等级(hospital level),\\
			医院类型(type of hospital), 性质(attribute), 占地面积(floor space), 隶属(affiliate), 行政区类别(type of administrative region)\\
			\hline
			\textbf{Class-Path: /物/药品/精神药品(/thing/drug/spirit drug)} \\ 
			\textbf{top-10 attributes: }药品类型(drug type), 用途分类(purpose), 规格(specification), 药品名称(name), 剂型(dosage form), \\
			用法用量(usage and dosage), 批准文号(license permission number), 别称(alias), 总部地点(headquarters location), 性质(nature)\\
			\hline
			\textbf{Class-Path: /物/生物/植物/树木/花树(/thing/biology/botany/tree/flower tree)}\\ 
			\textbf{top-10 attributes:} 门(phylum), 纲(class), 目(order), 属(genus), 界(kingdom),\\
			中文学名(Chinese scientific name), 拉丁学名(Latin scientific name), 科(family), 亚纲(subclass), 种(species)\\
			\hline
			\textbf{Class-Path: /人/名人/明星/亚洲明星(/person/famous person/star/Asia star)}\\ 
			\textbf{top-10 attributes:} 国籍(nationality), 出生地(birthplace), 出生日期(date of birth), 毕业院校(graduate institution), 职业(profession),\\
			主要成就(achievements), 民族(ethnicity), 性别(gender), 代表作品(representative work), 逝世日期(death date)\\
			\hline
		\end{tabular}
	\end{center}
	\caption{\label{tab: class2attribute}Cases of top-10 Predicted attributes of different domain Class-Path samples. }
\end{table*}

\section{Conclusion}

This paper investigates an automatic paradigm of \textit{attribute acquisition} in ontology
 by learning the continuous representation of class-paths and attributes, 
which is called \textbf{TransAtt}. 
To solve the problem of ambiguous allocating attributes to classes in data-driven knowledge bases,
TransAtt utilizes selective attention over class-paths.
Throughout the entire experiment, we construct a dataset called \textbf{BigCilin11K} and 
design two different attribute acquisition tasks to evaluate the performance of 
TransAtt and the learned distribution representation from different viewpoints and application context.
The results show that TransAtt has the ability of selecting over class-paths 
for attributes prediction of multi-roles entity,
and predicting attributes accurately for given class-paths even not seen before.
The mechanism of automatic attribute acquisition can help ontology construction achieve automatic attribute acquisition to 
further improve the scalability of ontology and efficiency of building a knowledge base.
Besides, though this work focus on Chinese ontology, 
our proposed method is language-independent for automatic attribute prediction.

\bibliographystyle{acl}
\bibliography{my.bib}

\end{CJK*}
\end{document}